\documentclass{article}

\date{}
\usepackage{arxiv}

\usepackage[utf8]{inputenc} %
\usepackage[T1]{fontenc}    %
\usepackage{hyperref}       %
\usepackage{url}            %
\usepackage{booktabs}       %
\usepackage{amsfonts}       %
\usepackage{nicefrac}       %
\usepackage{microtype}      %
\usepackage[pdftex]{graphicx}
\usepackage[table,xcdraw]{xcolor}

\usepackage{bm}
\usepackage{pifont}
\newcommand{\cmark}{\ding{51}}%
\newcommand{\xmark}{\ding{55}}%
\begin{document}

\title{K-VQG: Knowledge-aware Visual Question Generation for Common-sense Acquisition} %

\author{%
  Kohei Uehara \\
  The University of Tokyo\\
  \texttt{uehara@mi.t.u-tokyo.ac.jp} \\
\And
Tatsuya Harada \\
  The University of Tokyo / RIKEN\\
  \texttt{harada@mi.t.u-tokyo.ac.jp} \\
}

\maketitle

\begin{abstract}
Visual Question Generation (VQG) is a task to generate questions from images.
When humans ask questions about an image, their goal is often to acquire some new knowledge.
However, existing studies on VQG have mainly addressed question generation from answers or question categories, overlooking the objectives of knowledge acquisition.
To introduce a knowledge acquisition perspective into VQG, we constructed a novel knowledge-aware VQG dataset called K-VQG.
This is the first large, humanly annotated dataset in which questions regarding images are tied to structured knowledge.
We also developed a new VQG model that can encode and use knowledge as the target for a question.
The experiment results show that our model outperforms existing models on the K-VQG dataset.

\end{abstract}


\section{Introduction}\label{sec:introduction}
Asking questions is an important ability for humans in acquiring new knowledge.
Humans ask questions regarding what they see to acquire new knowledge and become more intelligent.
Therefore, to develop machine intelligence that can actively learn about the world, it is essential to study systems that can ask questions about what they see and acquire new knowledge.

Visual Question Generation (VQG) is a research field that aims to give machines such ability to ask questions about an image.
VQG was initially studied as a task that simply uses an image as input and generates a question related to the image~\cite{vqg}.
However, it is impossible to control the questions to be generated using only images as input because the targets and contents of the questions are extremely diverse.

\begin{figure}[t]
   \centering
   \includegraphics[width=0.75\linewidth]{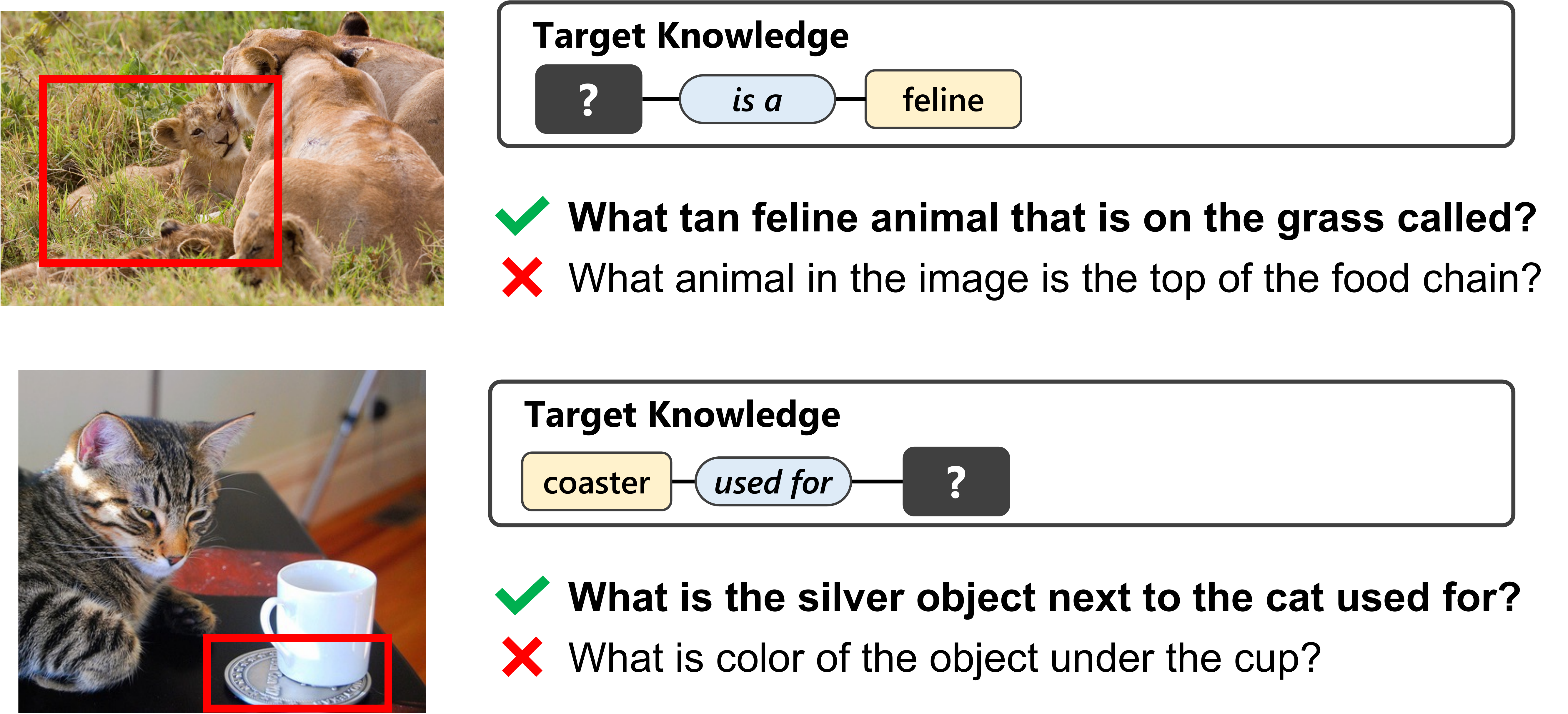}
   \caption{We proposed a new dataset and task in which the model is required to generate a target-knowledge aware question for a given image.
   In this task, the model is given a knowledge triplet with a missing part and is expected to generate a question that can complement the missing part.
   }
   \label{fig:intro}
\end{figure}

Recent research on VQG has focused on the way to providing information about the target of a question to the VQG model.
Existing studies have used possible answers~\cite{iqan,ivqa}, answer types~\cite{info-vqg,c3vqg}, answer categories~\cite{vqg_unknown}, and question-types~\cite{qt_vqg} as target information.
However, when using the answer as a condition, the answer to the question must be known before generating the question.
Since questions are usually asked without knowing the answer, such a problem setting is unnatural.
Other target information used in existing studies controls only the rough target of the question and cannot be used to generate questions that ask for specific knowledge.

To solve these problems and establish a more natural and practical setting for VQG, we introduce \textbf{K-VQG}, which is a task that utilizes \textit{target knowledge}, i.e., knowledge to be obtained by the question, as target information.
Following previous studies on structured knowledge~\cite{conceptnet,comet2020}, we represent knowledge as a triplet of three words or phrases, i.e., $<$head, relation, tail$>$.
Specifically, the model takes an image and \textit{masked target knowledge}, which is a knowledge triplet in which a part of the triplet is masked out as input and generates a question such that the answer will be helpful in complementing the missing part.

For example, in the top example of Figure~\ref{fig:intro}, the target knowledge is $<$lion, is a, feline$>$, and the masked target knowledge is $<$\textit{something}, is a, feline$>$.
The expected output would then be a question related to the knowledge and whose answer would be ``lion'', e.g., ``What tan feline animal that is on the grass called?''
On the other hand, the question like ``What animal in the image is the top of the food chain?'' is indeed a question whose answer will be ``lion'', but it is not related to the target knowledge.

Since there is no dataset with the necessary annotations (e.g., images, questions, and associated knowledge triplets) for this task, we constructed a new dataset called the \textbf{K-VQG dataset}.
Our K-VQG dataset is the first VQG dataset that is common-sense aware, human-annotated, and large-scale.

To solve this task, it is necessary to develop a model that can understand the visual information of an image and masked target knowledge information simultaneously.
Existing methods for VQG consider only simple target information, such as answers and categories, and thus cannot handle complex auxiliary information, such as a knowledge triplet.
Thus, we developed a novel model for K-VQG, which can encode the image and masked target knowledge using a multi-modal transformer based encoder to generate questions.

Our contributions are summarized as follows:
\begin{itemize}
    \item We introduce a novel VQG dataset with knowledge annotations called K-VQG.
    \item We propose a knowledge-aware VQG model that uses a masked knowledge triplet as input.
    \item We evaluate the performance of the proposed model on the constructed dataset.
\end{itemize}

\section{Related Work}\label{sec:related-works}

\subsection{Knowledge-aware VQA/VQG Dataset}\label{subsec:knowledge-aware-vqa/vqg-dataset}

In this section, we introduce Visual Question Answering (VQA) datasets in addition to VQG datasets because even datasets built for VQA can be used as VQG datasets by replacing the inputs and outputs.
We summarize the main features of various datasets in Table~\ref{tab:compare_dataset}.

The largest and best-known VQA dataset is the VQA v1/v2 dataset~\cite{vqa,vqav2}, which is also used in the VQA challenge competition.
The VQA v1/v2 dataset is the most commonly used dataset in VQG studies~\cite{ivqa,iqan,creativity,info-vqg}.
However, these datasets do not contain any knowledge annotations.

There are several knowledge-aware VQA datasets such as the FVQA~\cite{fvqa}, OK-VQA~\cite{okvqa}, K-VQA~\cite{kvqa}, and CRIC datasets~\cite{gao2019two}.
The FVQA dataset~\cite{fvqa}, in which the questions are annotated with common-sense triplets, is similar to our own.
However, the FVQA dataset is relatively small ($\sim$5K questions), and many of the questions tend to refer primarily to the target knowledge and less to the content of the images.
The questions in the FVQA dataset often refer to the image with only phrases like ``\ldots in the image''.
Such questions can be easily generated without understanding the content of the image and are therefore unsuitable for use in VQG.
The OK-VQA dataset~\cite{okvqa} is intended to be a VQA dataset that requires knowledge and is larger than the FVQA dataset ($\sim$10K questions); however, it lacks annotations on ``which knowledge is relevant to the question.''
The K-VQA dataset~\cite{kvqa} is specialized for knowledge of named entities (e.g., ``Who is to the left of Barack Obama?''), and its question annotations are template-based, making it less generalizable.
CRIC~\cite{gao2019two} is a more recently proposed dataset.
This dataset is similar to that proposed in our study in that it is a VQA dataset with common-sense triplet annotations.
However, this dataset is not annotated by humans, but is a rule-based dataset that automatically generates sentences from scene graph information.

Compared to the existing datasets mentioned above, our dataset is the first dataset that has all the features: the questions are associated with common-sense knowledge triplet, annotated by humans, bounding box annotations of the question target, and large scale.

\begin{table}[t]
\centering
\caption{
Comparison of key features of the major VQG/knowledge-aware VQA datasets.
Our dataset is the first manually-annotated VQG dataset that contains knowledge annotations and bounding box annotations.
}
\label{tab:compare_dataset}
\begin{tabular}{@{}lccccc@{}}
\toprule
 & Num. of Q & \begin{tabular}[c]{@{}c@{}}knowledge\\ type?\end{tabular} & \begin{tabular}[c]{@{}c@{}}structured\\ knowledge?\end{tabular} & \begin{tabular}[c]{@{}c@{}}target\\ bounding box?\end{tabular} & \begin{tabular}[c]{@{}c@{}}manually\\ annotated?\end{tabular} \\ \midrule
VQAv2~\cite{vqav2} & 1.1M & N/A & \xmark & \xmark & \cmark \\
FVQA~\cite{fvqa} & 5,826 & common-sense & \cmark & \xmark & \cmark \\
OK-VQA~\cite{okvqa} & 14,055 & open knowledge & \xmark & \xmark & \cmark \\
K-VQA~\cite{kvqa} & 183,007 & named entities & \cmark & \xmark & \xmark \\
CRIC~\cite{gao2019two} & 1.3M & common-sense & \cmark & \cmark & \xmark \\ \midrule[0.3pt]
\textbf{K-VQG} & \textbf{16,098} & \textbf{common-sense} & \textbf{\cmark} & \textbf{\cmark} & \textbf{\cmark} \\ \bottomrule
\end{tabular}
\end{table}

\subsection{VQG Model}\label{subsec:vqg-model}
VQG is the task of generating questions associated with images.
The earliest VQG model~\cite{vqg} used an RNN model to generate questions using only an image as the input.
However, such a model conditioned only on images cannot control the target of a question.
Therefore, researchers have been studying ways to control the target of a question by providing additional information.
In addition to images, iQAN~\cite{iqan} and iVQA~\cite{ivqa} use answers as inputs to generate questions that can produce the desired answers.
With these methods, the answer to the question must be known in advance.
Since questions are usually asked without knowing the answers, such problem setting is unnatural.

Other methods use categories of answers as conditions for VQG~\cite{info-vqg,c3vqg}.
With these methods, it is not necessary to know the answers to the questions; therefore, the aforementioned problem can be overcome.
However, there is a problem that the granularity of the answer categories greatly affect the quality of the control of the question content.
Although existing studies~\cite{info-vqg,c3vqg} use 15 categories, the classification is rather coarse because all answers related to the name of the object are gathered in the ``object'' category.
This means that, when there are multiple objects in an image, it is impossible to control which object should be the target of the generated question.

With our method, the input is a partially masked common-sense triplet.
Thus, our method has the advantage of being able to control the target in more detail than the existing VQG models, and it is also easy to apply the acquired information to a knowledge database.

\section{K-VQG Task and Dataset}\label{sec:k-vqg-task-and-dataset}
First, we provide an overview of the K-VQG task, which is a VQG task for knowledge acquisition.
In the K-VQG task, the model is given a \textbf{masked target knowledge} triplet and an \textbf{image}, and the model is expected to generate a question that can acquire the \textbf{target knowledge}.
The masked target triplet is a knowledge triplet in which a part of the question to be answered is masked, e.g., $<$[MASK], IsA, feline$>$.
By contrast, the target knowledge is a complete triplet in which the masked parts are filled, e.g., $<$lion, IsA, feline$>$.
For example, the goal of this task is to generate questions from a masked target triplet, such as $<$[MASK], IsA, feline$>$, such that ``lion'' can be obtained as an answer, and knowledge $<$lion, IsA, feline$>$ can be acquired.

Next, we describe the construction of the K-VQG dataset.
Each sample in the dataset contains the following information: the (1) image, (2) question, (3) answer, (4) target knowledge triplet, (5) bounding box of the question target.

We asked crowd workers of Amazon Mechanical Turk (AMT)\footnote{https://www.mturk.com/} to annotate the data.
We sampled the images from the Visual Genome dataset~\cite{visualgenome} and selected the target object and candidates for the target knowledge (Subsection~\ref{subsubsec:common-sense-triplet-collection.} (a)).
We then asked the workers to select one target knowledge and write questions about the image that required the target knowledge to answer (Subsection~\ref{subsubsec:question-collection.} (b)).
We further conduct the question validation process to ensure the quality of the dataset (Subsection~\ref{subsubsec:question-validation.} (c)).

\subsection{Dataset Construction}\label{subsec:dataset-construction}

\subsubsection{(a) Knowledge triplet collection.}\label{subsubsec:common-sense-triplet-collection.}

We utilized ConceptNet~\cite{conceptnet} and \textsc{Atomic$^{20}_{20}$}~\cite{comet2020} as the sources of the common-sense triplets.

ConceptNet is a large-scale knowledge base that contains knowledge collected from several resources.
Knowledge in ConceptNet is represented as a triplet of the form $<$head, relation, tail$>$, such as $<$cat, AtLocation, sofa$>$.
ConceptNet contains approximately 34 million triplets and 37 types of relations.
Some relations seem to be unnatural targets for questions regarding images, such as \textit{DistinctFrom} or \textit{MotivatedByGoal}.
Thus, we selected 15 types of relations that were considered suitable as targets for the questions.

\begin{figure}[t]
    \centering
    \includegraphics[height=12.0cm]{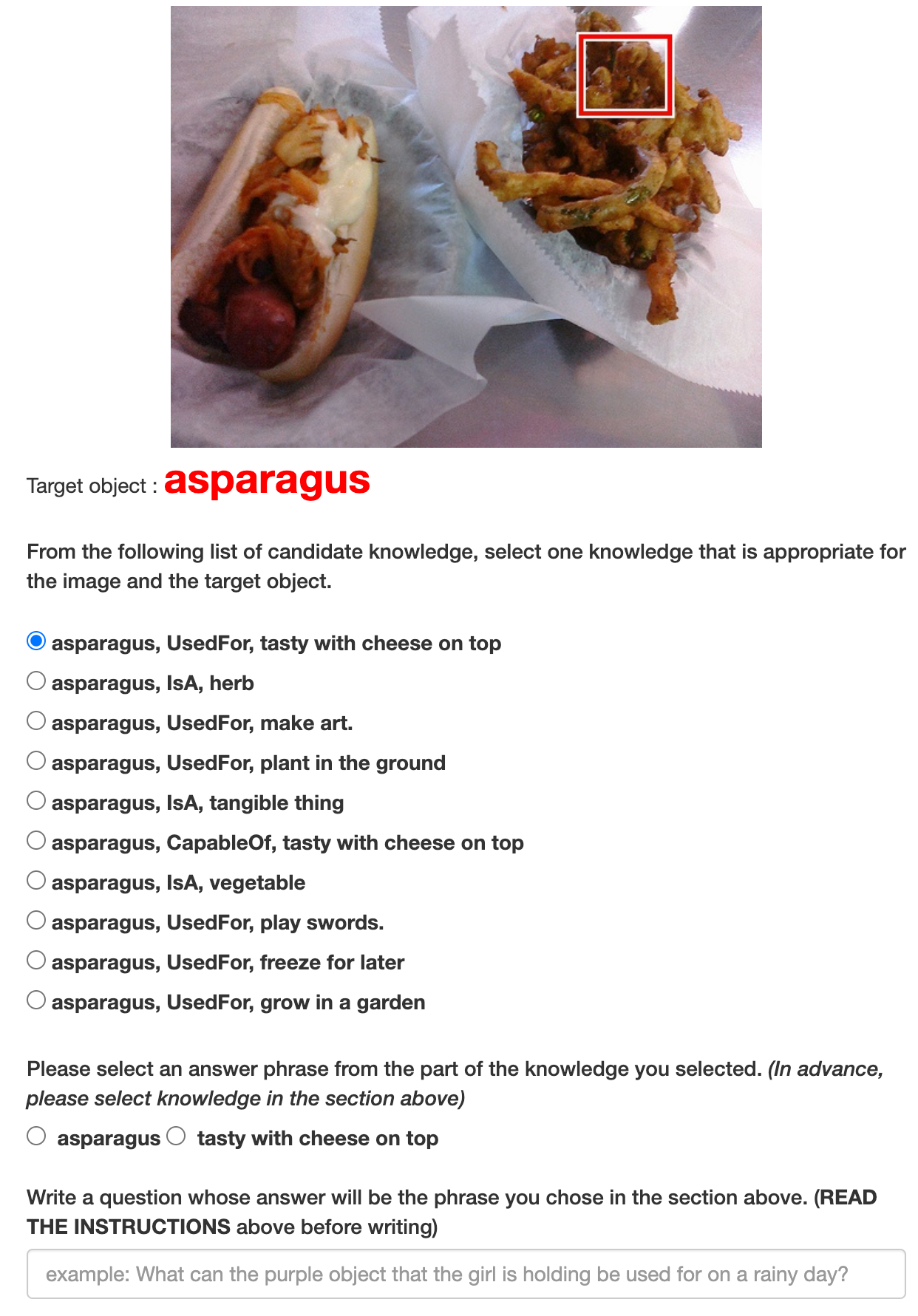}
    \caption{
        Screenshot of the AMT task (excluding instruction due to space limitation).
    The information provided to the worker was displayed at the top of the screen, including the image, target object, and candidate knowledge triplets.
   Below that, there are sections for the selection of the answer phrases and writing knowledge-aware questions corresponding to the selected answers.
    }
    \label{fig:screenshot}
\end{figure}

The second source of knowledge is \textsc{Atomic$^{20}_{20}$}.
The \textsc{Atomic$^{20}_{20}$} consists of more then 1M knowledge triplets about physical-entity relations (e.g., $<$bread, ObjectUse, make french toast$>$), event-centered relations (e.g., $<$PersonX eats spinach, isAfter, PersonX makes dinner$>$), and social-interactions (e.g., $<$PersonX calls a friend, xIntent, to socialize with their friend$>$).
We used only physical-entity relations for our dataset construction because the other relation types were less relevant to the images in the Visual Genome.

After the above pre-processing, we merged these two knowledge datasets.
Then, to remove knowledge that is unrelated to any objects in the images, we queried the entity appearing as the head of the knowledge in the Visual Genome object list and removed the knowledge if there was no matching object.

Finally, we obtained a total of $\sim$150K knowledge triplets as candidate knowledge.

\subsubsection{(b) Question collection.}\label{subsubsec:question-collection.}

We show the screenshot of the AMT interface in Figure~\ref{fig:screenshot}.

In order to maintain quality, we selected workers who resided in the U.S. or Canada and had an approval rate greater than 97\%.
The workers were given the following information: the target image, a bounding box representing the area of the target object (i.e., the head entity of the candidate knowledge), the name of the target object, and a list of candidate knowledge triplets (up to 15).
They were then asked to write knowledge-aware questions with the following steps:
\begin{enumerate}
    \item From the list of candidate knowledge, select one knowledge that is appropriate for the image and the target object.
    \item Select a phrase from the selected knowledge (i.e., head entity or tail entity) to be the answer.
    \item Write a question whose answer will be the selected phrase and requires the knowledge the worker have chosen to answer.
\end{enumerate}

\setlength{\tabcolsep}{0.75em}
\begin{table}[]
\centering
\caption{\textbf{Dataset Statistics.}
We compare the K-VQG dataset with FVQA and VQA v2 dataset.
\textit{Num. of head/tail answers} indicate the number of answers which is the head or tail entity of the knowledge triplet.
Note that the FVQA dataset does not provide such information, and we automatically counted the number.
However, because of spelling inconsistencies, we could not obtain an exact count, and thus we used an approximate number here.
}
\label{tab:question-statistics}
\begin{tabular}{@{}lccc@{}}
\toprule
 & \textbf{K-VQG} & FVQA~\cite{fvqa} & VQAv2~\cite{vqav2} \\ \midrule
Num. of questions & 16,098 & 5,826 & 443,757 \\
-- Num. of head answers & 11,588 & $\sim$4,430 & N/A \\
-- Num. of tail answers & 4,510 & $\sim$1,240 & N/A \\
Num. of images & 13,648 & 2,190 & 82,783 \\ \midrule[0.3pt]
Num. of unique answers & 2,819 & 1,427 & 22,531 \\
Num. of unique knowledge & 6,084 & 4,180 & N/A \\ \midrule[0.3pt]
Num. of unique head & 527 & 847 & N/A \\
Num. of unique tail & 4,922 & 2,871 & N/A \\\midrule[0.3pt]
Average answer length & 1.46 & 1.23 & 1.10 \\
Average question length & 13.88 & 9.55 & 6.20 \\ \midrule[0.3pt]
Num. of non-knowledge words in questions & 3.35 & 0.99 & N/A \\ \bottomrule
\end{tabular}
\end{table}

To make the VQG model capable of properly understanding the content of an image, it is desirable for the questions to describe the relationships between objects in the image.
Thus, we instructed the workers to write questions that included a description of the position of the object in relation to other objects in the image, more than simple phrases such as ``\ldots in the image''.
In addition, we instructed them to assume that the bounding box of the target object is not visible, i.e.,  phrases such as ``surrounded by a red frame'' or ``with a bounding box'' are prohibited.

\subsubsection{(c) Question validation.}\label{subsubsec:question-validation.}

To ensure the quality of the collected questions, we further conducted validation of the collected annotations by AMT.
We asked workers to evaluate questions with the following criteria: (1) whether the question refers to the visual content of the image, (2) whether the target knowledge is related to the question, (3) whether the target knowledge is related to the image and the target object, (4) whether the question contains typos or grammatical errors, (5) whether the answer is proper for the question.
We asked three workers per question for evaluation, and excluded the questions in which all workers unanimously gave negative ratings for any of the evaluation criteria.
Note that we evaluated some of the data ourselves in advance, and rejected submissions from workers whose agreement rate with our evaluation was less than 60\%, in order to maintain the quality of the evaluation.

\subsection{Dataset Statistics}\label{subsec:data-statistics}

\begin{figure}[t]
\centering
\includegraphics[width=1.0\linewidth]{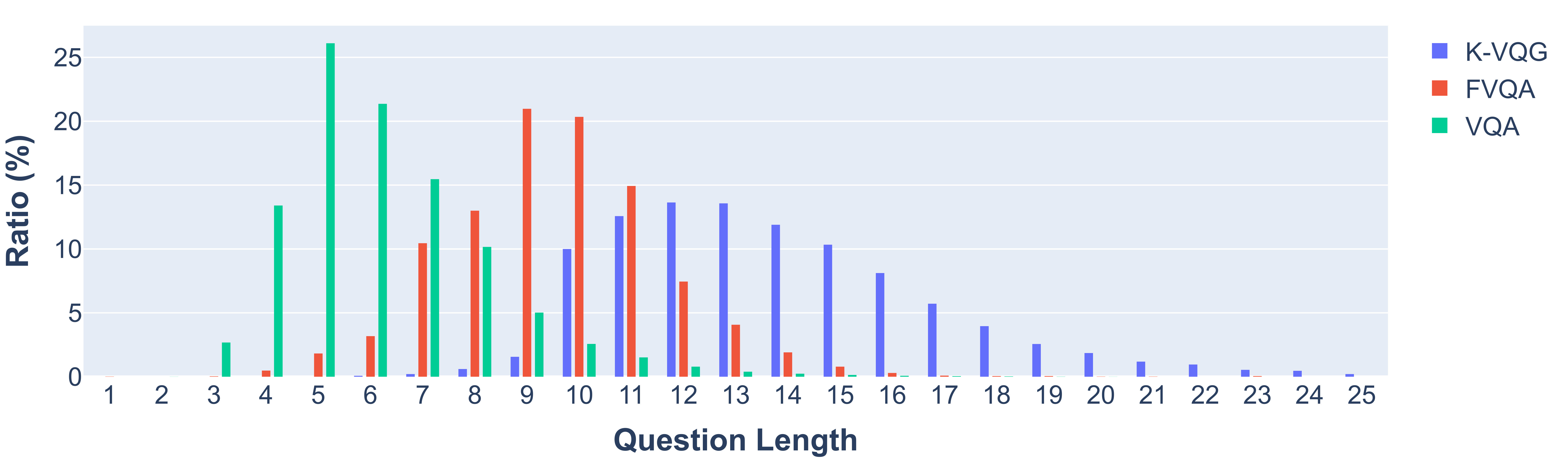}
\caption{The distribution of question lengths in K-VQG dataset, FVQA dataset and VQA v2 dataset.
The K-VQG dataset tends to have longer questions than the other datasets.
}
\label{fig:q_length}
\end{figure}

\begin{figure}[t]
\centering
\includegraphics[width=1.0\linewidth]{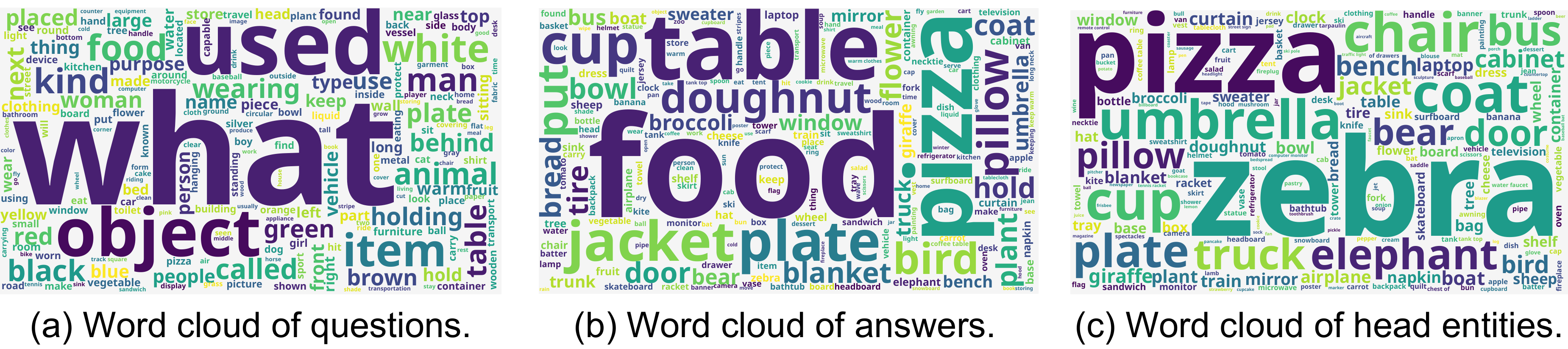}
\caption{
    Word cloud for questions, answers, and head entities in the K-VQG dataset.
Note that basic stopwords are excluded for the word cloud for questions.
}
\label{fig:wordcloud}
\end{figure}

\begin{figure}[t]
\centering
\includegraphics[width=0.9\linewidth]{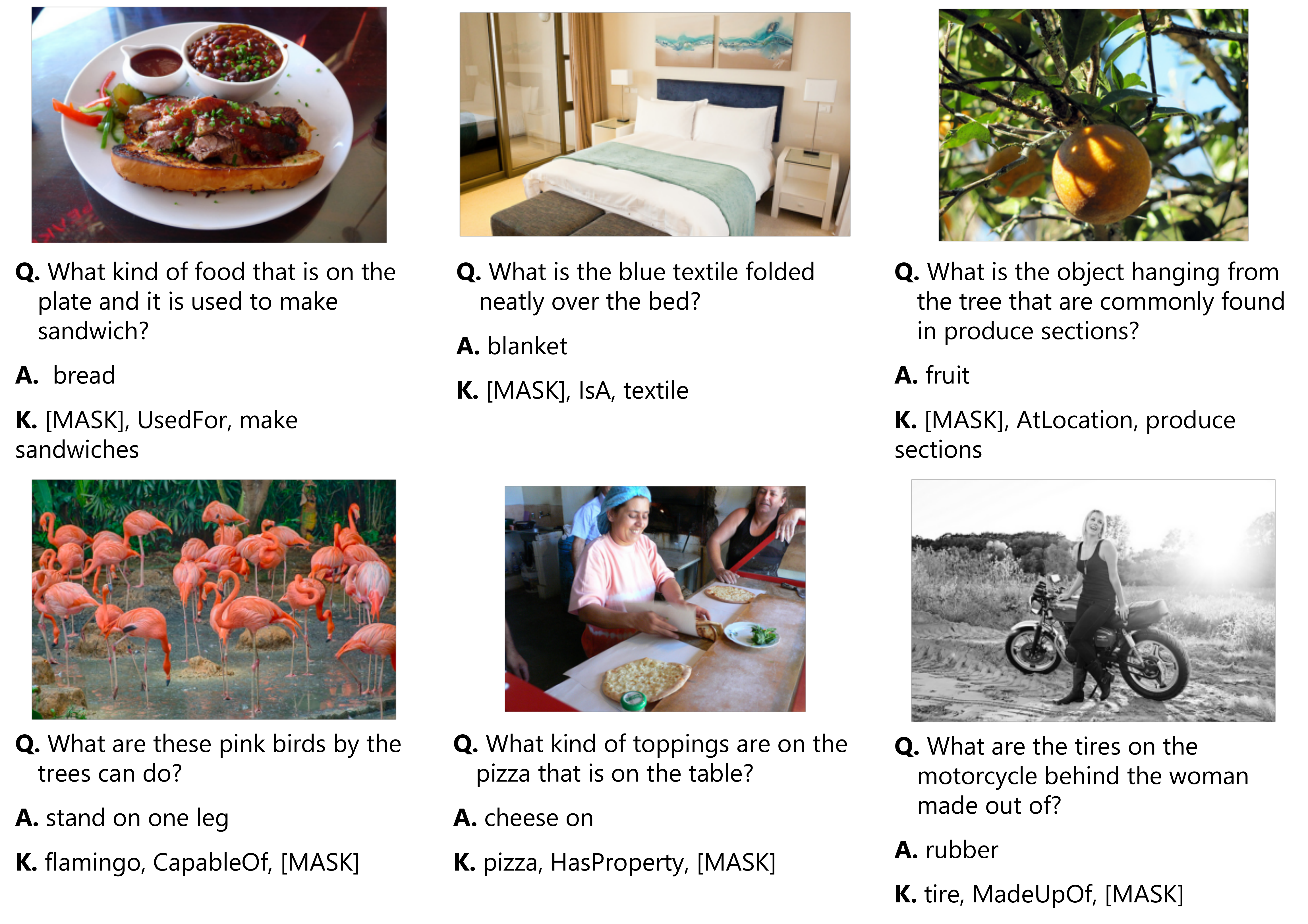}
\caption{Example questions and the corresponding images, answers, target knowledge from the K-VQG dataset.}
\label{fig:examples}
\end{figure}

The basic statistics of our dataset and two existing datasets, FVQA and VQAv2, are shown in Table~\ref{tab:question-statistics}.

We collected 16,098 questions, corresponding to 13,648 images, and 6,084 knowledge triplets.
The K-VQG dataset has 2,819 unique answers.
Among the 5,220 knowledge triplets, there are 527 unique heads, 15 unique relations, and 4,922 unique tails.
Our K-VQG dataset is significantly larger than FVQA, which is an existing knowledge-aware dataset.

From these statistics, we can conclude that our dataset is more challenging for the VQG model.
First, our dataset has a longer sentence length for questions and answers than other datasets.
This can also be observed in Figure~\ref{fig:q_length}, which shows the distribution of the length of the questions in each dataset.
In addition, compared with FVQA, our dataset has more non-knowledge words in the questions (3.35 vs. 0.99).
This means that most of the questions in FVQA consist of words derived from the knowledge triplet, whereas the questions in our dataset contain many words that are not derived from the knowledge triplet (e.g., words regarding the content of the image).
In other words, for the FVQA dataset, the VQG model could generate questions without understanding the content of the images.
However, because the questions in the K-VQG dataset require references to the image content, generating proper questions is much more difficult.

In Figure~\ref{fig:wordcloud}, we show the word clouds of the most frequent words in the questions, answers, and head entities in the annotations of the K-VQG dataset.
This indicates that the questions in our dataset pertain to diverse fields (e.g., food, animals, and location).

We show some examples from the K-VQG dataset in Figure~\ref{fig:examples}.

\section{Model}\label{sec:k-vqg-model}

\begin{figure}[t]
   \centering
   \includegraphics[width=0.9\linewidth]{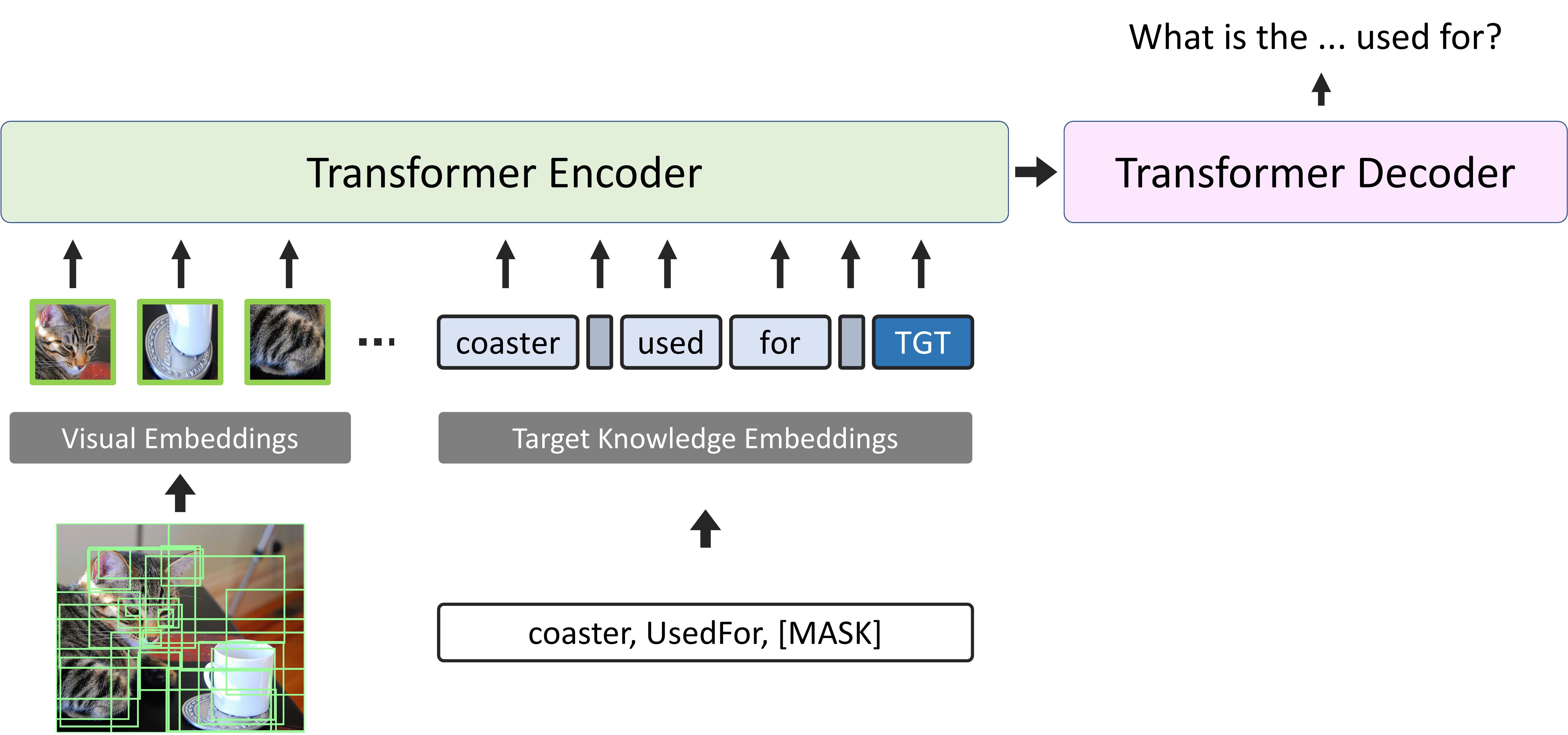}
   \caption{The overview of the model.
   Our model takes an image and a target knowledge triplet as input, and convert them to fused features by multi-modal Transformer encoder.
   Then, a Transformer decoder takes the fused features as input and generates a question in an auto-regressive manner.
   }
   \label{fig:model}
\end{figure}

We developed a baseline question generation model, which is an encoder-decoder type model using a transformer~\cite{transformer}.
The overview of the model is shown in Figure~\ref{fig:model}.
Our model consists of an encoder that encodes the image information and target knowledge information and a decoder that uses the output of the encoder to generate the question.

\subsection{Encoder}\label{subsec:encoder}
Our encoder uses a pre-trained UNITER~\cite{uniter}, which is a multi-modal transformer model that can encode image and text information.
In this model, the encoder takes the visual embeddings $\bm{v}$ of the image and the target knowledge embeddings $\bm{k}$ as the input and outputs the encoded representation $\bm{h}$, that is, $\bm{h} = \mathrm{Enc}(\bm{v},\;\bm{k})$.

\subsubsection{Visual Embeddings.}\label{subsubsec:visual-embeddings}
To obtain visual embeddings $\bm{v}$, we use a pre-trained Faster R-CNN model~\cite{faster} and extract region features~\cite{butd} of the image.
Following ~\cite{uniter}, to provide the positional information of each image region, a seven-dimensional vector representing the coordinates and area of the region was encoded by a linear layer and added to the region image features.

\subsubsection{Target Knowledge Embeddings.}\label{subsubsec:target-knowledge-embeddings.}
As described in Section~\ref{sec:k-vqg-task-and-dataset}, we used partially masked knowledge triplets as input to the model.
We treat the masked target knowledge triplet as a sequence of words.
Input masked target knowledge is tokenized as asequence of tokens $\bm{k} = \{\bm{w}_h,\;w_{\mathtt{[SEP]}},\;\bm{w}_r,\;w_{\mathtt{[SEP]}},\;\bm{w}_{t}\}$.
Here, $w_{\mathtt{[SEP]}}$ is a special token that indicates the separation of each part, and $\bm{w}_h,\;\bm{w}_r,\;\bm{w}_t$ denote the tokens of the head, relation, and tail phrases, respectively, e.g., $\bm{w}_h = \{w_{h1},\;w_{h2},\;\ldots,\;w_{hn}\}$.
If the head or tail is the masked part, token $\bm{w}$ is replaced by a special token $w_{\mathtt{[TGT]}}$.

\subsection{Decoder}\label{subsec:decoder}
The decoder is a module that receives the encoded input image and target knowledge, and outputs the question, that is, $\bm{q} = \mathrm{Dec}(\bm{h})$.
Following the recent success of transformers in language generation, we developed a transformer-based model for the decoder.
Our decoder is an autoregressive transformer model, adapted from BART~\cite{bart}, and consists of several transformer blocks, each of which has a multi-head cross-attention and self-attention mechanism.

Our model was trained in a teacher-forcing manner by minimizing the negative conditional log-likelihood loss.
The loss function can be expressed through the following equation:

\begin{equation}
    L_{LM} = -\sum^{|\bm{q}|}_{n=1}\log P_{\theta}(\bm{q}_n\,|\,\bm{q}_{<n},\,\bm{h}).\label{eq:equation}
\end{equation}

\subsection{Implementation Details}\label{subsec:implementation-details}
Following UNITER, we set the number of Transformer blocks in the encoder and decoder to 12, and the number of hidden units in each block to 768.
We initialized the weights of the encoder from the pre-trained UNITER model\footnote{downloaded using the script at \url{https://github.com/ChenRocks/UNITER}}.
We used the AdamW optimizer~\cite{adamw} with $\beta_1 = 0.9$ and $\beta_2 = 0.999$.
As the learning rate scheduling, we adapted the cosine annealing scheduling, where warm-up steps were set to 10\% of the total training steps.
The maximum learning rate was set to be $1.0 \times 10^{-5}$.
We trained the model for 2K steps.
The training took two hours on 8 $\times$ Tesla A100 GPU.

\section{Experiments}\label{sec:experiments}
We tested our model and several existing methods on the K-VQG dataset.
We split the dataset into training and validation sets, and we used the validation set to evaluate the performance of the model.
Out of total of 16,098 questions, 12,891 questions were used for training, and 3,207 questions were used for validation.
Note that we made sure to split the dataset so that the images used in the UNITER pre-training did not contaminate the validation split of the dataset.

\subsection{Baselines.}\label{subsec:baselines.}
We used several existing methods as the baselines.
We did not use any answer-aware VQG models because we did not assume a situation in which the model already knew the expected answer.
Thus, we pick VQG models that take images and/or answer categories as input.
We automatically annotated the answer categories.
If the answer is a word that is the head of the knowledge triplet, we use hypernym dictionary in WordNet~\cite{wordnet} to determine the answer category.
If the answer is the tail of the triplet, we use this relation as the answer category.

\begin{description}
    \item \textbf{I2Q}~\cite{vqg}: The I2Q model is a baseline model based on the approach in ~\cite{vqg} that uses only the image as the input and generates a question.
    \item \textbf{IC2Q}: The IC2Q model  uses the image and the answer category as the inputs.
    \item \textbf{V-IC2Q}~\cite{creativity,info-vqg}: The V-IC2Q model is a variational auto-encoder (VAE) based method, which encodes the answer category and question into a latent space, and decodes the latent vector to generate a question.
    \item \textbf{IM-VQG}~\cite{info-vqg}: IM-VQG model is another VAE based method. The model is trained to maximize the mutual information between the image, question, and expected answer. Simultaneously, another latent space is learned to encode the answer category, which enables the model to generate questions from only the image and category inputs, without any expected answers.
\end{description}

\subsection{Input ablation.}\label{subsec:input-ablation.}
To demonstrate the importance of input to the model, we performed an input ablation study in which either the image or the target knowledge is excluded from the input to the model (\textbf{Ours w/o image}, \textbf{Ours w/o knowledge}).

\subsection{Evaluation metrics.}\label{subsec:evaluation-metrics}
Following previous VQG research, we used \textbf{BLEU}~\cite{bleu}, \textbf{METEOR}~\cite{meteor}, and \textbf{CIDEr}~\cite{cider} as evaluation metrics.

In the K-VQG task, it is also important to evaluate whether the generated questions  correctly yield the target knowledge.
To this end, we used the Target Knowledge Parser to predict the masked target knowledge triplet from the generated questions and checked the consistency with the expected knowledge triplet.
The Target Knowledge Parser has a similar structure as the K-VQG model.
It has a UNITER-based encoder to encode images and questions and a BART-based decoder to generate/recover masked target knowledge.
We used \textbf{Triplet-BLEU} to evaluate the overall agreement between the generated triplets and the ground truth by calculating the BLEU-4 score.
In addition, we used \textbf{Head-Acc}, \textbf{Relation-Acc}, and \textbf{Tail-Acc} to evaluate whether each part of the triplet is correct.

\setlength{\tabcolsep}{5pt}
\begin{table}[t]
\centering
\caption{Qualitative results on the K-VQG dataset.
The left-side of the table shows the metrics used to evaluate the quality of the questions.
Here, B-4, M, and C represent BLEU-4, METEOR, and CIDEr, respectively.
The right-side of the table shows the metrics for the knowledge consistency.
Tri-BLEU, H-Acc, R-Acc, and T-Acc denote Triplet-BLEU, Head-Acc, Relation-Acc, and Tail-Acc, respectively.
For all metrics, higher values are better.}
\label{tab:results}
\begin{tabular}{@{}l
>{\columncolor[HTML]{FFE7E6}}c
>{\columncolor[HTML]{FFE7E6}}c
>{\columncolor[HTML]{FFE7E6}}c
>{\columncolor[HTML]{FFFED3}}c
>{\columncolor[HTML]{FFFED3}}c
>{\columncolor[HTML]{FFFED3}}c
>{\columncolor[HTML]{FFFED3}}c @{}}
\toprule
 & \multicolumn{3}{c}{\cellcolor[HTML]{FFE7E6}Question Quality} & \multicolumn{4}{c}{\cellcolor[HTML]{FFFED3}Knowledge Consistency} \\
 & B-4 & M & C & Tri-BLEU & H-Acc & R-Acc & T-Acc \\ \midrule
I2Q~\cite{vqg} & 11.74 & 17.05 & 27.30 & 4.50 & 69.69 & 55.35 & 1.15 \\
IC2Q & 12.37 & 16.69 & 31.01 & 7.97 & 75.34 & 58.62 & 27.91 \\
V-IC2Q~\cite{creativity,info-vqg} & 11.78 & 17.18 & 28.72 & 4.70 & 68.66 & 55.60 & 1.53 \\
IM-VQG~\cite{info-vqg} & 11.44 & 17.07 & 26.19 & 4.10 & 68.07 & 55.32 & 1.71 \\ \midrule[0.3pt]
Ours w/o image & 17.28 & 21.06 & 113.1 & 61.99 & 81.95 & 83.13 & 58.59 \\
Ours w/o knowledge & 10.65 & 16.45 & 33.92 & 6.99 & 65.73 & 51.01 & 4.37 \\
\textbf{Ours} & \textbf{18.84} & \textbf{22.79} & \textbf{131.04} & \textbf{64.33} & \textbf{84.72} & \textbf{82.44} & \textbf{66.20} \\ \bottomrule
\end{tabular}
\end{table}

\subsection{Results}\label{subsec:results}
We show the experimental results in Table~\ref{tab:results}.
The left side of the table shows the results in terms of the quality of the generated questions, and the right side shows the metric of whether the generated questions yield the desired knowledge.

\subsubsection{Question Quality (vs. baselines)}
For all metrics used to evaluate the quality of the question, our method outperformed the baselines (Ours vs. others).
The baseline method uses only image (I2Q) or image and category (IC2Q, V-IC2Q, IM-VQG) information as input for inference, which suggests that the model has not achieved the ability to sufficiently control the content of the questions to be generated.
By contrast, our method directly encodes the target knowledge information and thus succeeds in generating questions with content closer to the ground truth.

\begin{figure}[!t]
   \centering
   \includegraphics[width=0.95\linewidth]{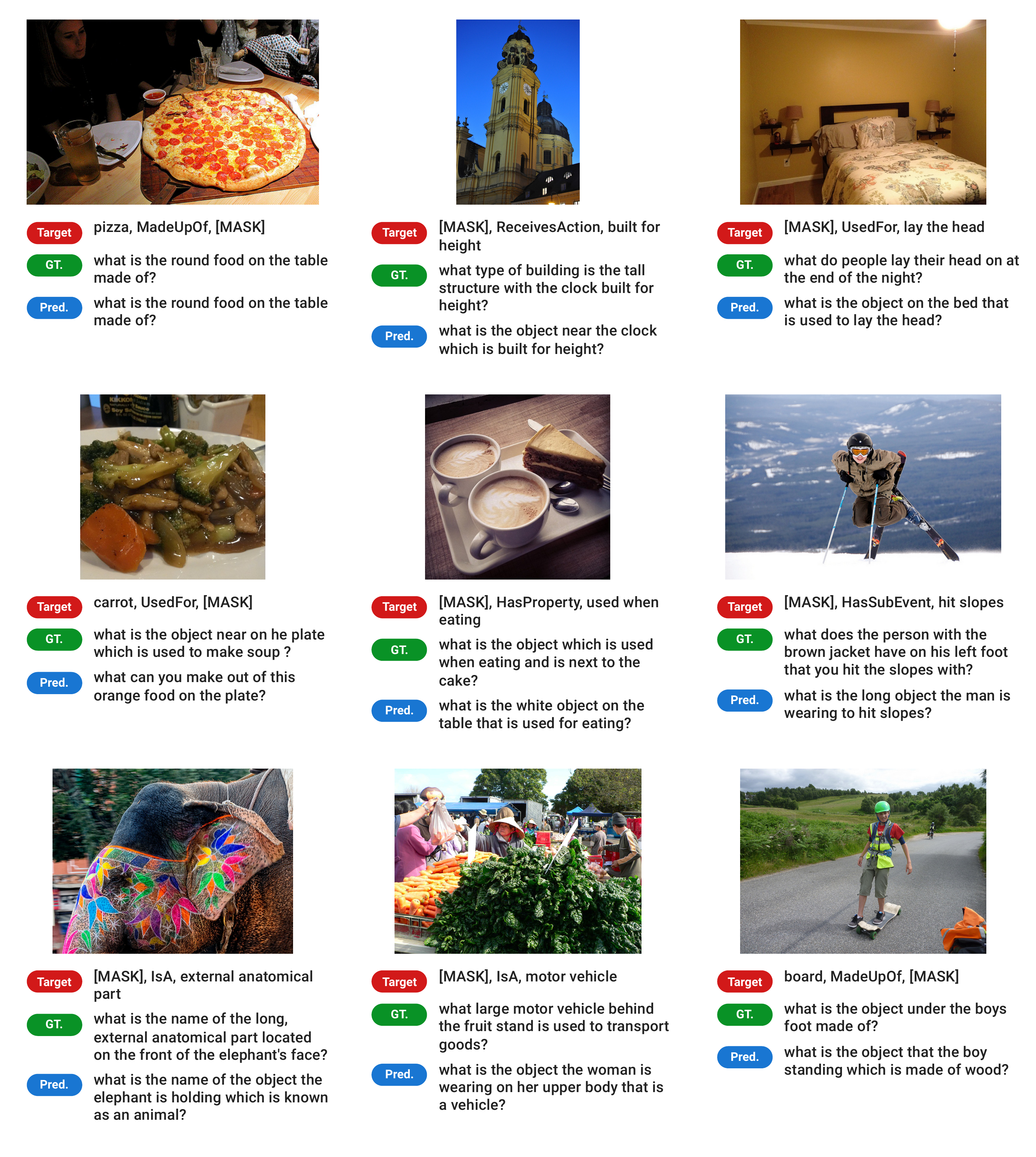}
   \caption{
       Output examples of our method on the K-VQG dataset.
   We show the input images, target knowledge, ground-truth questions, and generated questions.
   }
   \label{fig:qualitative}
\end{figure}

\subsubsection{Knowledge Consistency (vs. baselines)}
The right side of Table~\ref{tab:results} shows the metrics for knowledge consistency.
In terms of Tri-BLEU, which evaluates the overall quality of the generated triplet, our method significantly improves the score compared with other methods.
In addition, for part-level accuracy (Head-Accuracy, Relation-Accuracy, and Tail-Accuracy), our method outperformed the other methods.
For Head-Accuracy and Relation-Accuracy, our method outperformed the other methods, but the difference was smaller than with the Tail-Accuracy.
This is likely due to the fact that the head and relation are often shorter and less diverse than the tail, making it relatively easy to answer correctly even with a conventional method.
It should be noted that although tails consist of multiple words, which makes it difficult to generate them correctly, our method can achieve a fairly high accuracy.

\subsubsection{Input Ablation.}
From the input ablation study, it can be seen that when only one of the inputs (image or target knowledge) is used, the performance is worse than when both are used.
The performance degradation is particularly noticeable when no target knowledge is input.
This may be because target knowledge contains more information about question content control than images.
That is, when target knowledge is input, information about what the answer should be is available to the model, whereas when only images are input, such information critical to question content control is not available.

These results highlights our claim that the use of desired knowledge as input is important for controlling the content of VQG.

\subsubsection{Output Examples.}
We show several examples of generated questions in Figure~\ref{fig:qualitative}.
In general, our method successfully generates questions that capture the input target knowledge and the content of the images.
The bottom three are examples where our model failed to output.
From these failed examples, we can see that our model sometimes fails to generate questions when the target object is hidden or too small.
In the case of the bottom-right example, the generated question is indeed related to the target knowledge, but the question is about the board itself, not the board material.
We believe that further research in methods of encoding image content and knowledge targets will lead to more precise control of question generation.

\section{Conclusion}\label{sec:conclusion}
In this study, we introduce a novel VQG task that uses knowledge as the target of the question.
To this end, we constructed a novel knowledge-aware VQG dataset called the K-VQG dataset.
The K-VQG dataset is the first large-scale and manually annotated knowledge-aware VQG dataset.

We also developed a benchmark model for the K-VQG task.
Our experiments demonstrated the effectiveness of our method, while showing some room for improvement.

For future research, our proposed task and dataset have a variety of potential applications.
Given the nature of the task, in which the model acquires new knowledge by asking questions, we believe that this task can contribute to the development of learning frameworks, such as human-in-the-loop and learning-by-asking~\cite{lba}.
We expect that this research will lead to the development of a proactive learning system that acquires information about the external world as images and actively learns new knowledge from humans by asking them questions about the images.

\section*{Acknowledgements}\label{sec:acknowledgements}
This work was partially supported by JST AIP Acceleration Research JPMJCR20U3, Moonshot R\&D Grant Number JPMJPS2011, JSPS KAKENHI Grant Number JP19H01115, and JP20H05556 and Basic Research Grant (Super AI) of Institute for AI and Beyond of the University of Tokyo.
We would like to thank Naoyuki Gunji, Qier Meng for the helpful discussions.

\bibliographystyle{splncs04}
\bibliography{egbib}

\clearpage
\appendix
\section{Appendix}

\subsection{Details of Target Knowledge Parser}\label{sec:details-of-target-knowledge-parser}
Following K-VQG model, we used a model that consists of a UNITER-based encoder~\cite{uniter} and BART-based decoder~\cite{bart} as our Target Knowledge Parser model.
The encoder takes the visual embeddings $\bm{v}$ and the tokenized question $\bm{q}$.
We used region features obtained from Faster R-CNN~\cite{butd} as visual embeddings, as in our VQG model.
The question is tokenized into input sequences using WordPiece tokenizer~\cite{wordpiece}.

Our model is trained to minimizing the negative conditional log-likelihood loss function can be expressed through the following equation:

\begin{equation}
    L = -\sum^{|\bm{k}|}_{n=1}\log P_{\theta}(\bm{k}_n\,|\,\bm{k}_{<n},\,\bm{h}_t)\label{eq:triplet_loss}
\end{equation}
where $\bm{h}_t = \textrm{Enc}(\bm{v},\;\bm{q})$, and $\bm{k} = \{\bm{w}_h,\;w_{\mathtt{[SEP]}},\;\bm{w}_r,\;w_{\mathtt{[SEP]}},\;\bm{w}_{t}\}$.
 is a special token that indicates the separation of each part, and $\bm{w}_h,\;\bm{w}_r,\;\bm{w}_t,\;w_{\mathtt{[SEP]}}$ denote the tokens of the head, relation, tail phrases and special token, respectively.

\subsection{Additional Examples of the K-VQG Dataset}\label{sec:additional-examples-of-the-k-vqg-dataset}
We show additional examples of the K-VQG dataset below.

\vspace*{\fill}
\begin{figure}[h]
   \centering
   \includegraphics[width=1.0\linewidth]{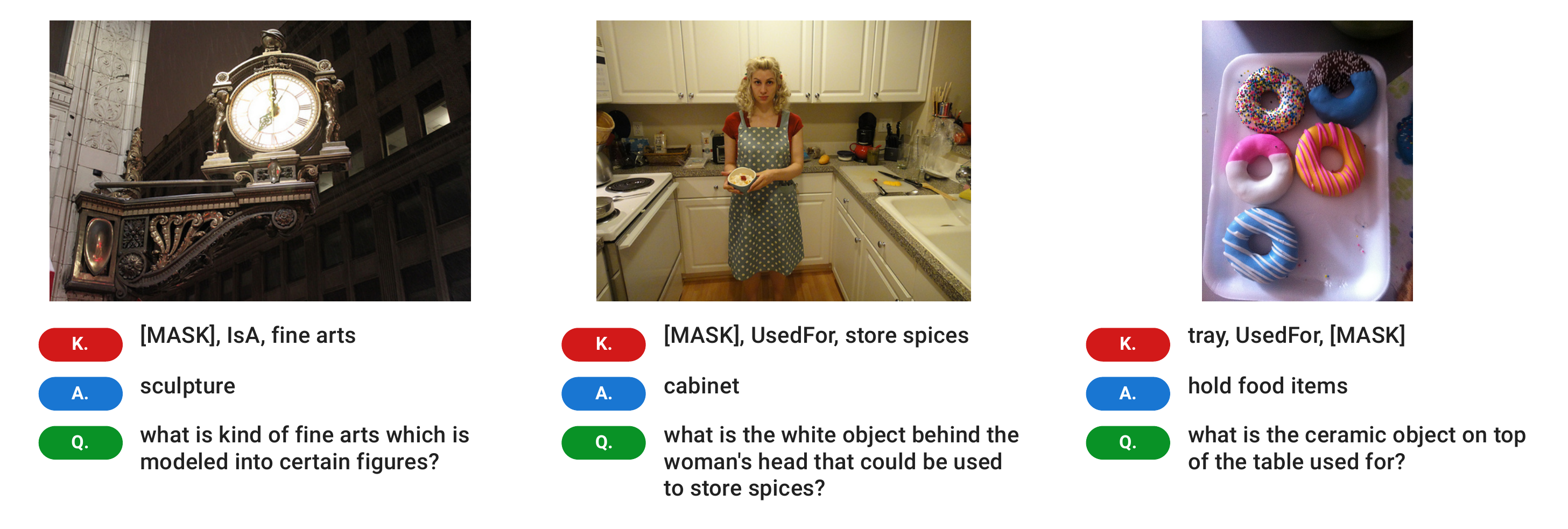}
   \caption{
Additional examples of the K-VQG dataset (1)
   }
   \label{fig:examples0}
\end{figure}
\vspace*{\fill}

\begin{figure}[!t]
   \centering
   \includegraphics[width=1.0\linewidth]{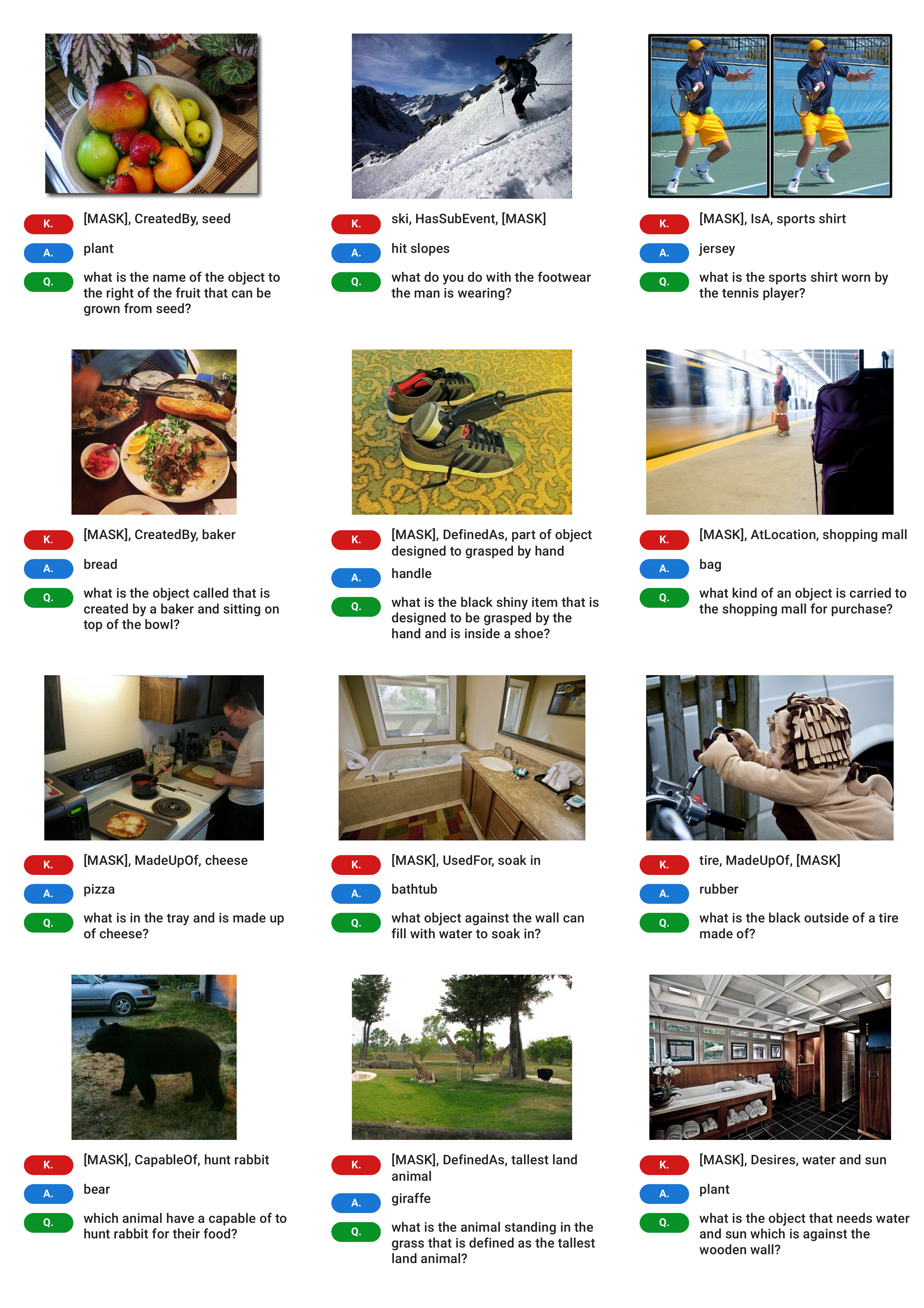}
   \caption{
Additional examples of the K-VQG dataset (2)
   }
   \label{fig:examples1}
\end{figure}

\begin{figure}[!t]
   \centering
   \includegraphics[width=1.0\linewidth]{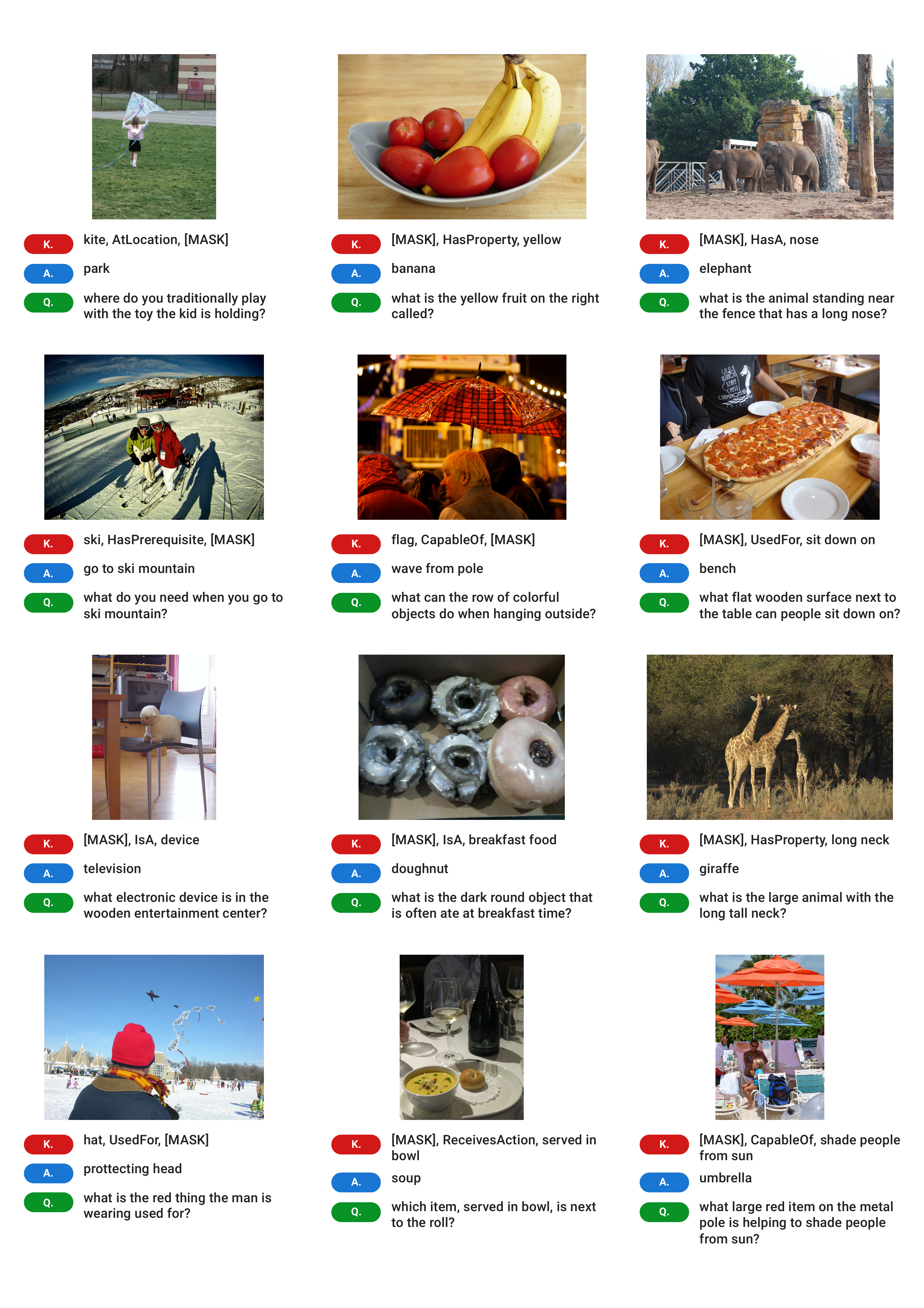}
   \caption{
Additional examples of the K-VQG dataset (3)
   }
   \label{fig:examples2}
\end{figure}

\end{document}